\documentclass[lettersize,journal]{IEEEtran}
\usepackage{amsmath,amsfonts}
\usepackage{algorithmic}
\usepackage{algorithm}
\usepackage{array}
\usepackage[caption=false,font=normalsize,labelfont=sf,textfont=sf]{subfig}
\usepackage{textcomp}
\usepackage{stfloats}
\usepackage{url}
\usepackage{verbatim}
\usepackage{graphicx}
\usepackage{cite}
\usepackage{tikz}
\usepackage{comment}
\usepackage{amsmath,amssymb} 
\usepackage{color}
\usepackage{booktabs}  
\usepackage{threeparttable} 
\usepackage{multicol}  
\usepackage{multirow}  

\usepackage[normalem]{ulem}
\useunder{\uline}{\ul}{}
\usepackage{float}

\usepackage{tabularx}
\usepackage{longtable}
\usepackage[colorlinks,linkcolor=black]{hyperref}

\begin{document}

\title{Cross-Dataset-Robust Method for Blind Real-World Image Quality Assessment}

\author{Yuan Chen, Zhiliang Ma and Yang Zhao

\thanks{
Y. Chen is with the School of Internet, Anhui University, Hefei 230039, China (e-mail: ychen@ahu.edu.cn).

Z. Ma and Y. Zhao are with the School of Computers and Information, Hefei University of Technology, Hefei 230009, China (e-mail: mzl@mail.hfut.edu.cn; yzhao@hfut.edu.cn).

Y. Zhao is the corresponding author.

}
}



\maketitle

\begin{abstract}
Although many effective models and real-world datasets have been presented for blind image quality assessment (BIQA), recent BIQA models usually tend to fit specific training set. Hence, it is still difficult to accurately and robustly measure the visual quality of an arbitrary real-world image. In this paper, a robust BIQA method, is designed based on three aspects, i.e., robust training strategy, large-scale real-world dataset, and powerful backbone. First, many individual models based on popular and state-of-the-art (SOTA) Swin-Transformer (SwinT) are trained on different real-world BIQA datasets respectively. Then, these biased SwinT-based models are jointly used to generate pseudo-labels, which adopts the probability of relative quality of two random images instead of fixed quality score. A large-scale real-world image dataset with 1,000,000 image pairs and pseudo-labels is then proposed for training the final cross-dataset-robust model. Experimental results on cross-dataset tests show that the performance of the proposed method is even better than some SOTA methods that are directly trained on these datasets, thus verifying the robustness and generalization of our method. 
\end{abstract}


\section{Introduction}
\IEEEPARstart{I}{mage} quality assessment (IQA) is a basic topic in the field of computer vision and image processing, which tends to evaluate the perceptual quality of images like human beings. For decades, objective IQA methods have attracted a lot of attention. According to the usage of reference images, IQA algorithms can be categorized into full-reference (FR) \cite{Wang2004ImageQA, Zhang2011FSIMAF}, reduced-reference (RR) \cite{Wang2018ReducedReferenceQA, Yu2022PerceptualHW} and no-reference (NR) methods. No-reference IQA, also known as blind IQA (BIQA), is commonly used in practical applications because of the lack of references in many real-world scenes.

Owing to the powerful learning ability for complex nonlinear regression problems, many deep neural network (DNN)-based models \cite{Kang2014ConvolutionalNN,Zhang2020BlindIQ,Pan2022DACNNBI,Song2022BlindIQ,Ying2020FromPT,Gao2023BlindIQ} have been introduced to BIQA task in recent years. However, these deep models usually contain millions of parameters, and require large numbers of labeled samples for training. Therefore, many effective BIQA datasets with mean opinion score (MOS) labels have been proposed, such as LIVEC  \cite{Ghadiyaram2016MassiveOC}, BID\cite{Ciancio2011NoReferenceBA}, KonIQ-10k \cite{Hosu2020KonIQ10kAE}, SPAQ \cite{Fang2020PerceptualQA}, and FLIVE \cite{Ying2020FromPT}. Unfortunately, obtaining reliable subjective quality scores is an extremely time-consuming and tedious process, so the scale and diversity of the existing IQA databases are still
\begin{figure}
\centering
\includegraphics[width=1.0\columnwidth]{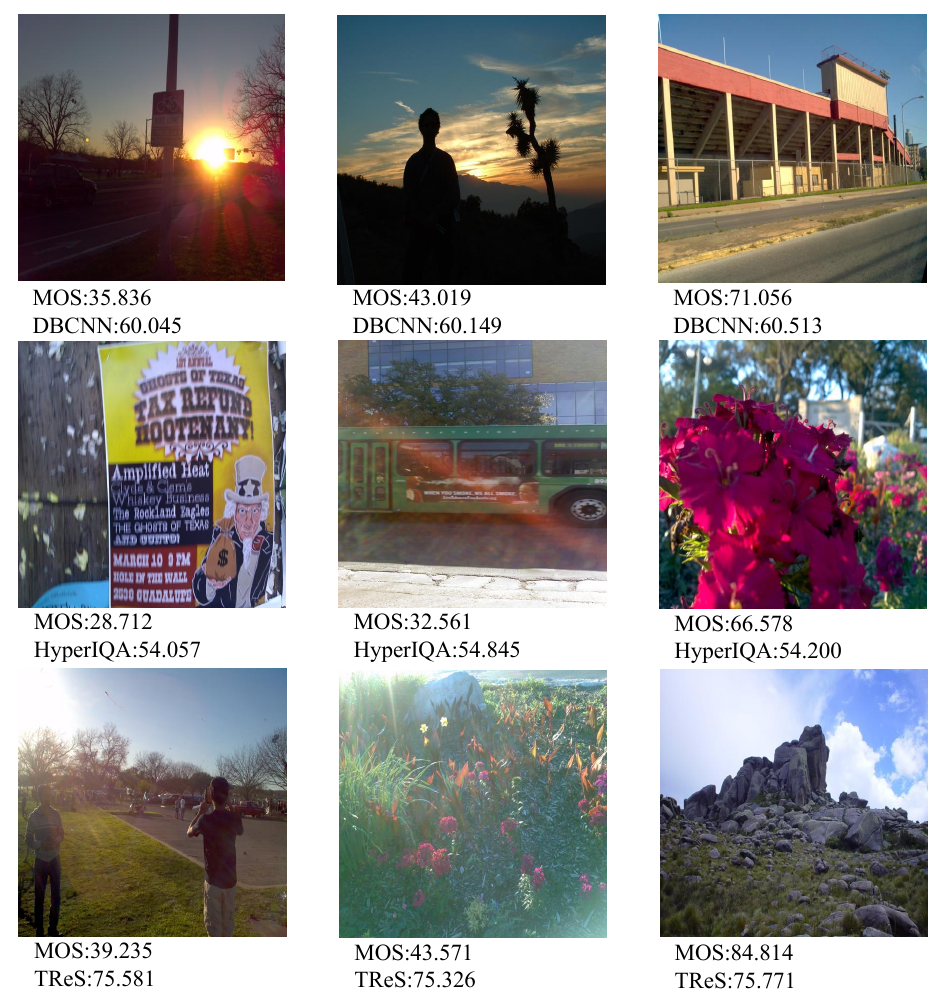}
\caption{Cross-dataset test results of DBCNN \cite{Zhang2020BlindIQ}, HyperIQA \cite{Su2020BlindlyAI} and TReS \cite{Golestaneh2022NoReferenceIQ}. DBCNN, HyperIQA and TReS are all trained on KonIQ-10k \cite{Hosu2020KonIQ10kAE}, and the test images are selected from LIVEC \cite{Ghadiyaram2016MassiveOC}.}
\label{fig1}
\end{figure}
insufficient. In addition, MOS labels of different datasets are produced with different observers, evaluation criteria and experimental environments. Obviously, it is impossible to enumerate each situation of infinite real-world images. Therefore, theoretically speaking, these real-world BIQA datasets are still biased. If a DNN-based model with powerful learning capability can achieve best results on a dataset by training on itself, this model may become more biased due to overfitting to a specific training set. As a result, the generalization and robustness of many SOTA BIQA networks may still not meet practical demands.

As shown in Fig. \ref{fig1}, three typical deep BIQA models trained on KonIQ-10k dataset, i.e., DBCNN \cite{Zhang2020BlindIQ}, HyperIQA \cite{Su2020BlindlyAI} and TReS \cite{Golestaneh2022NoReferenceIQ}, may produce inconsistent results to MOS labels on another LIVEC dataset. More cross-dataset test results are listed in Table \ref{tab1}. We can find that the networks can achieve best scores on the training set, but their performance drops significantly on other datasets. These results demonstrate that although the existing BIQA datasets contain as many images as possible and subjective MOS labels as accurate as possible, DNNs still tend to overfit the MOS of specific training set, so they cannot generate a reliable score for an arbitrary real-world image out of the training set.

In order to break through the bottleneck of large-scale dataset with subjective labels, recent BIQA algorithms have explored various strategies. For example, there are several methods \cite{Ma2017dipIQBI},\cite{Lin2020DeepFLIQAWS} that adopt FR-IQA scores as labels on large dataset with known degradations. However, FR-IQA labels are generated with fixed degradations and still be different to human visual perception. To reduce the bias of each dataset, some methods \cite{Zhang2021UncertaintyAwareBI,
Zhang2020LearningTB,Sun2021BlindQA} mix several different datasets for training, but directly mixed data with different distributions may lead to more complicated optimization. Therefore, pseudo-labels based on relative probability are adopted in \cite{Wang2021LearningFS},\cite{Wang2021SemiSupervisedDE}, which are much easier to obtain than large-scale subjective annotations. 

In this paper, a cross-dataset-robust BIQA (CDR-BIQA) model is designed, which mainly adopts three strategies, i.e., more robust training strategy, better network backbone, and larger real-world image set. Firstly, SOTA Swin-Transformer (SwinT) \cite{Liu2021SwinTH} is used as backbone to train individual SwinT-IQA model on each BIQA dataset, respectively. Then, these SwinT-IQA models with bias are used to estimate relative probability instead of absolute scores, which is used as pseudo-labels on a larger image set in the wild. Finally, the final CDR-BIQA model is trained with this large-scale real-world dataset and pseudo-labels. The contributions of this paper are summarized as follows,

\begin{table}[tp]
\centering  
\renewcommand\arraystretch{1.2}
\fontsize{6.1}{7}\selectfont 
\caption{SRCC and PLCC evaluations on cross dataset tests.}
\begin{tabular}{cccccccc}
\toprule[1pt]
                                 & \multirow{3}{*}{\begin{tabular}[c]{@{}c@{}}Training\\ Set\end{tabular}} & \multicolumn{6}{c}{Testing Set}                                                                     \\ \cline{3-8} 
                                 &                                                                         & \multicolumn{2}{c}{LIVEC}       & \multicolumn{2}{c}{BID}         & \multicolumn{2}{c}{KonIQ-10k}   \\ \cline{3-8} 
                                 &                                                                         & SRCC           & PLCC           & SRCC           & PLCC           & SRCC           & PLCC           \\ \toprule[1pt]
\multirow{3}{*}{DBCNN\cite{Zhang2020BlindIQ}}    & LIVEC                                                                   & \textbf{0.851} & \textbf{0.869} & 0.809          & 0.832          & 0.754          & 0.825          \\
                                 & BID                                                                     & 0.751          & 0.814          & \textbf{0.845} & \textbf{0.859} & 0.734          & 0.795          \\
                                 & KonIQ-10k                                                               & 0.755          & 0.777          & 0.815          & 0.818          & \textbf{0.875} & \textbf{0.884} \\ \toprule[1pt]
\multirow{3}{*}{HyperIQA\cite{Su2020BlindlyAI}} & LIVEC                                                                   & \textbf{0.859} & \textbf{0.882} & 0.756          & 0.790          & 0.747          & 0.808          \\
                                 & BID                                                                     & 0.790          & 0.840          & \textbf{0.869} & \textbf{0.878} & 0.717          & 0.785          \\
                                 & KonIQ-10k                                                               & 0.767          & 0.788          & 0.797          & 0.799          & \textbf{0.906} & \textbf{0.917} \\ \toprule[1pt]
\multirow{3}{*}{TReS\cite{Golestaneh2022NoReferenceIQ}}    & LIVEC                                                                   & \textbf{0.846} & \textbf{0.877} & 0.809          & 0.793          & 0.779          & 0.815          \\
                                 & BID                                                                     & 0.810          & 0.822          & \textbf{0.855} & \textbf{0.871} & 0.806          & 0.805          \\
                                 & KonIQ-10k                                                               & 0.734          & 0.806          & 0.742          & 0.791          & \textbf{0.915} & \textbf{0.928} \\ \toprule[1pt]

\end{tabular}
\label{tab1}
\end{table}

\begin{enumerate}
    \item A robust training strategy is proposed. Multiple BIQA models are firstly trained on several current datasets. Subsequently, these biased models are jointly used to determine the image quality on a large-scale real-world image dataset. A total of 500,000 to 1,000,000 pairs of samples are randomly selected from the image dataset, and the relative probability that two images are relatively good or bad is used to optimize final CDR-BIQA model instead of MOS labels. 
    \item Owing to the pre-trained SwinT backbone, each SwinT-IQA model can have powerful learning capability. Through training and testing on each BIQA dataset with MOS, it can be verified that simple usage of SwinT backbone can outperform many SOTA BIQA models.
    \item Experimental results of cross-dataset testing demonstrate the generalization ability of the proposed method. Cross-dataset testing results of the proposed method can be even better than some SOTA methods directly trained on these datasets.
  \end{enumerate}

\begin{figure*}
\centering
\includegraphics[height=9cm]{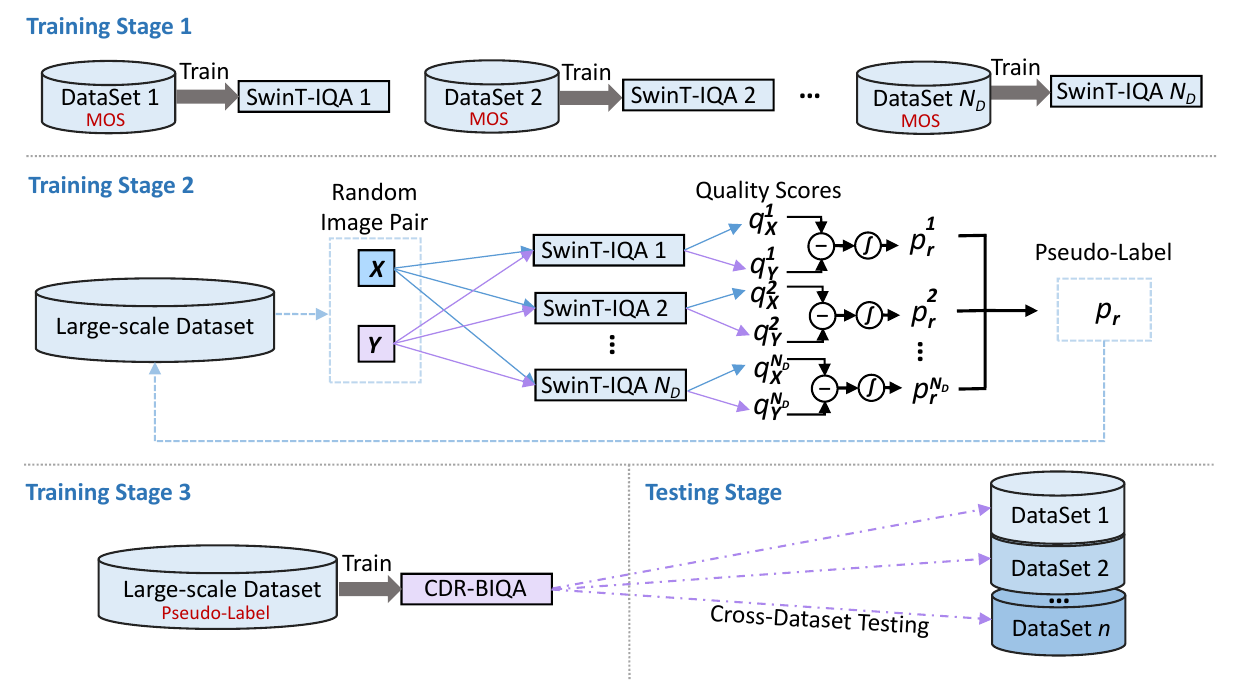}
\caption{Training strategy of the proposed method based on large-scale dataset and pseudo-labels.}
\label{fig2}
\end{figure*}

\section{Related Work}
\subsection{Blind Image Quality Assessment}
According to different features extracted for BIQA, it can be divided into two categories: based on hand-crafted feature and based on learning feature. Natural Scene Statistics (NSS) is a hand-crafted feature commonly used in BIQA models. Since high-quality natural scene images obey certain statistical properties, distortions will cause images to deviate from these statistics. The level of distortions in quality can thus be detected and quantified by modeling scene statistics which is sensitive to the appearance of distortion. These quality-aware natural scene parameters include discrete wavelet coefficients \cite{Moorthy2010ATF}, DCT coefficients \cite{Saad2012BlindIQ}, the correlation coefficients across subbands \cite{Moorthy2011BlindIQ}, locally normalized luminance coefficients with their pairwise products \cite{Mittal2012NoReferenceIQ}, image gradient, log-Gabor responses and color statistics \cite{Zhang2015AFC}. These hand-crafted features, however, require expertly design and are time-consuming. Furthermore, since the scene statistics represent image quality from a global view, the quality cannot be effectively evaluated for locally distorted in-the-wild images.

Inspired by the successes of machine learning in many computer vision tasks, some learning-based approaches are proposed. In the early stage, some codebook-based learning approaches are introduced \cite{Zhang2014TrainingQF,Ye2012NoReferenceIQ,
Xu2016BlindIQ,Ye2012UnsupervisedFL}. In recent years, deep convolutional neural networks(CNN) have shown strong learning capabilities, and thus the use of deep CNN for blind image quality assessment has become a trend. Kang \textit{et al}. \cite{Kang2014ConvolutionalNN} used a shallow CNN model composed of a convolutional layer and two fully connected layers to estimate the quality of small image patches, and then averaged the image-level quality scores with the predicted patch scores. Bosse \textit{et al}. \cite{Bosse2018DeepNN} further deepened the CNN model by jointly learning the quality and weight of each patch, where the weight is the relative importance of the patch quality for the global quality estimation.
 Liu \textit{et al}. \cite{Liu2017RankIQALF} first trained a Siamese Network to learn the quality level of a pair of images, and then fine-tuned the trained Siamese Network on the existing IQA dataset. Ma \textit{et al}. \cite{Ma2018EndtoEndBI} proposed a multi-task deep neural network based on joint recognition of distortion types and prediction of image quality. Zhang \textit{et al}. \cite{Zhang2020BlindIQ} proposed a model composed of two streams of deep CNNs, specializing in synthetic and authentic distortions scenarios separately. Su \textit{et al}. \cite{Su2020BlindlyAI} developed an adaptive hyper network to aggregate local distortion features and global semantic features. You \textit{et al}. \cite{You2021TransformerFI}  proposed an architecture in which a shallow transformer encoder is used on the feature maps extracted by a convolutional neural network. Ke \textit{et al}. \cite{Ke2021MUSIQMI} proposed a multi-scale image quality transformer, which uses the transformer architecture to solve image problems of different sizes and aspect ratios. Golestaneh \textit{et al}. \cite{Golestaneh2022NoReferenceIQ} proposed a combination of features extracted based on CNNs and Transformers, and a relative ranking loss that takes into account the relative ranking information between images. Sun \textit{et al}. \cite{Sun2021BlindQA} proposed an iterative mixed database training strategy, which can train BIQA models on multiple databases at the same time.
\subsection{Pseudo-Labels for BIQA}
Due to insufficient training data, some studies have tried to learn quality perception features from the pseudo-labels IQA dataset, where the pseudo-labels IQA dataset can be labeled with ease. Some studies uses the quality scores calculated by the most advanced FR-IQAs as pseudo-labels. Ma \textit{et al}. \cite{Ma2017dipIQBI} used RankNet to learn an opinion-unaware BIQA from millions of image pairs with identifiable quality. Wang \textit{et al}. \cite{Wang2021LearningFS} used domain adaptive method to solve the problem of mismatch between synthetic distortion and authentic distortion. Wu \textit{et al}. \cite{Wu2020EndtoEndBI} used a large-scale IQA dataset with pseudo-labels to pretrain the IQA model. However, most of them are developed on synthetic degraded images and thus perform poorly on the real-world IQA dataset, which is demonstrated in Section \uppercase\expandafter{\romannumeral4}.

\subsection{Uncertainty-Aware BIQA}

Learning uncertainty helps to understand and analyze model predictions. In the context of BIQA, Huang \textit{et al}. \cite{Huang2019ConvolutionalNN} modeled the uncertainty of patch quality to alleviate the label noise problem in patch-based training. Wu \textit{et al}. \cite{Wu2018BlindIQ} used a sparse Gaussian process for quality regression, in which the uncertainty of the data can be jointly estimated without supervision. To propose an effective BIQA model with probabilistic interpretation, Zhang \textit{et al}. \cite{Zhang2021UncertaintyAwareBI},\cite{Zhang2020LearningTB} used a hypothetical Thurstone model to create probabilistic labels through MOS and variance of each pair of images and learn the uncertainty of the data under direct supervision.

\section{The Proposed Method}
\subsection{Robust BIQA Learning Strategy}
In order to train a robust BIQA model, the most straightforward way is to build a large-scale dataset with accurate subjective labels. However, as mentioned before, it is extremely time-consuming and labor-intensive to obtain large-scale and reliable MOS labels. Therefore, this paper tends to design the following strategies to improve robustness in the training phase. First,  large-scale real-world images are collected for training. Increasing the number of real-world training samples is an important way to improve the performance and robustness of deep BIQA models. Second, motivated by \cite{Zhang2021UncertaintyAwareBI},\cite{Zhang2020LearningTB},\cite{Wang2021LearningFS}, relative probability values are used as labels instead of fixed MOS values. Compared to fit fixed values, relative probability can reduce the bias in different datasets. Third, several BIQA models trained on different datasets are jointly used to generate pseudo-labels. A single BIQA model also inherits the bias of data distribution on specific dataset. Motivated by classic assemble learning theory, such as random forest, the combination of these biased individual predictors can reproduce more robust results.

The detailed training framework is shown in Fig. \ref{fig2}. Firstly, a total of $N_{D}$ SwinT-IQA networks are trained on current BIQA datasets with MOS labels, respectively. Note that the MOS labels of different datasets are all linearly rescaled to [0, 1]. Secondly, the trained SwinT-IQA models are jointly used to produce pseudo-labels for the unlabeled large-scale real-world dataset. For each random image pair $(\boldsymbol{X}, \boldsymbol{Y})$ , these pre-trained SwinT-IQA models are individually applied to evaluate the quality score of image $\boldsymbol{X}$ and image $\boldsymbol{Y}$.  Let $q_{\boldsymbol{X}}^{i}$ and $q_{\boldsymbol{Y}}^{i}$ represent the estimated quality scores of $\boldsymbol{X}$ and $\boldsymbol{Y}$ by using the \textit{i-th} SwinT-IQA model. Then the relative probability $p_{r}^{i}(\boldsymbol{X}, \boldsymbol{Y})$ is computed for the \textit{i-th} SwinT-IQA model to take place of quality score values, as follows,

\begin{equation}
p_{r}^{i}(\boldsymbol{X}, \boldsymbol{Y})=f_{SIG}\left(q_{\boldsymbol{X}}^{i}-q_{\boldsymbol{Y}}^{i}\right)=\frac{1}{1+e^{\left(q_{Y}^{i}-q_{X}^{i}\right)}} ,
\end{equation}
where $f_{SIG}$ denotes the normalization operation, which adopts a sigmoid function instead of linear normalization in this paper. If the quality difference between $\boldsymbol{X}$ and $\boldsymbol{Y}$ is significant and easy to distinguish, the network need not pay too much attention to studying this situation. Conversely, if the quality levels of the two images are very similar, it is not reasonable to forcibly compare their visual quality, because $\boldsymbol{X}$ and $\boldsymbol{Y}$ have different contents. Therefore, the sigmoid-based normalization is used in this paper to restrain these two situations, and enforce the network to learn other situations finely.

After computing $p_{r}^{i}(\boldsymbol{X}, \boldsymbol{Y})$ of each SwinT-IQA model, the corresponding pseudo-label  $p_{r}(\boldsymbol{X}, \boldsymbol{Y})$ of image pair $(\boldsymbol{X}, \boldsymbol{Y})$ is calculated as,
\begin{equation}
p_{r}(\boldsymbol{X}, \boldsymbol{Y})=\frac{1}{N_{D}} \sum_{i=1}^{N_{D}} p_{r}^{i}(\boldsymbol{X}, \boldsymbol{Y}).
\end{equation}

In this way, we can produce labels $p_{r}(\boldsymbol{X}, \boldsymbol{Y})$ for the large-scale real-world image dataset. In the third stage, the final CDR-BIQA model, which utilizes the same SwinT-IQA structure, is trained on this large-scale dataset.

After training stages, the final CDR-BIQA model is directly tested on different BIQA datasets, without retraining or fine-tuning on them. Current BIQA methods are usually trained and tested on the same dataset. This way can verify the learning ability of proposed method. However, it cannot well test the generalization capability. The cross-dataset test is more helpful to verify the robustness of the proposed BIQA method.

\begin{figure*}
\centering
\includegraphics[height=3.55cm]{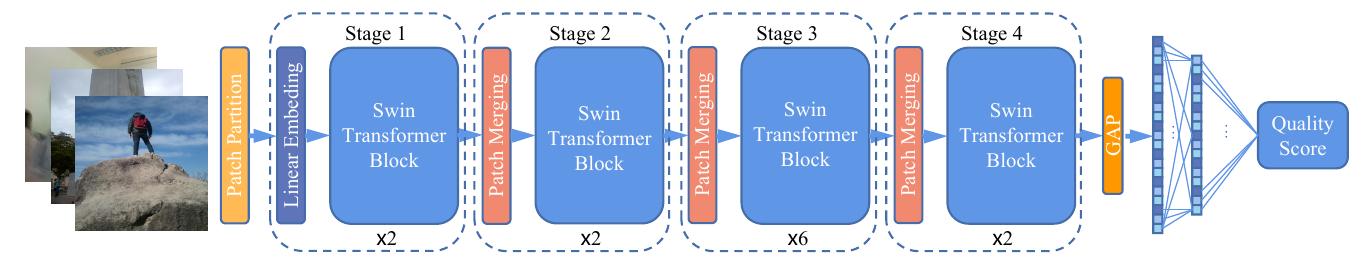}
\caption{Network architecture of the SwinT-IQA model. The model includes a Swin-Transformer-Tiny\cite{Liu2021SwinTH} for quality-aware feature extraction and an image quality regressor for mapping quality-aware features into the quality score space.}
\label{fig3}
\end{figure*}

\subsection{Structure of the SwinT-IQA}
	In order to evaluate the effectiveness of the proposed learning strategy, we didn't introduce too complicated network structure. The architecture of SwinT-IQA is shown in Fig.\ref{fig3}, which consists of a SwinT backbone and an image quality score regression module. Transformer structure \cite{Vaswani2017AttentionIA} was first proposed in natural language processing field. Subsequently, it was also applied to the field of computer vision and has achieved excellent performance \cite{Liu2021SwinTH},\cite{Dosovitskiy2021AnII} in many tasks. In order to take advantage of the powerful learning ability of transformer structure, this paper tends to use a pre-trained SwinT as the backbone. 

After extracting quality perception features through the effective SwinT backbone, a regression model is added to map these features to the quality score. The global average feature pooling (GAP) is firstly applied to generate a feature vector with dimension $\boldsymbol{P}$×1, where $\boldsymbol{P}$ dentoes the number of final feature maps. Then two fully connected (FC) layers are used to map the feature vector to the predicted quality score. In this paper, the two FC layers are composed of 512 neurons and 1 neuron, respectively. Finally, we can train the SwinT-IQA on current BIQA datasets in an end-to-end training manner with $\mathcal{L}_{1}$ loss function,

\begin{equation}
\mathcal{L}_{1}=\frac{1}{N} \sum_{i=1}^{N}\left\|q_{\text {score }}-q_{\text {label }}\right\|_{1} ,
\end{equation}
where $q_{score}$ and $q_{label}$ denote the predicted score and MOS of the \textit{i-th} training patch, and $N$ represents the total number of training patches.

\subsection{Details of the CDR-BIQA}
\begin{figure*}
\centering
\includegraphics[height=7.3cm]{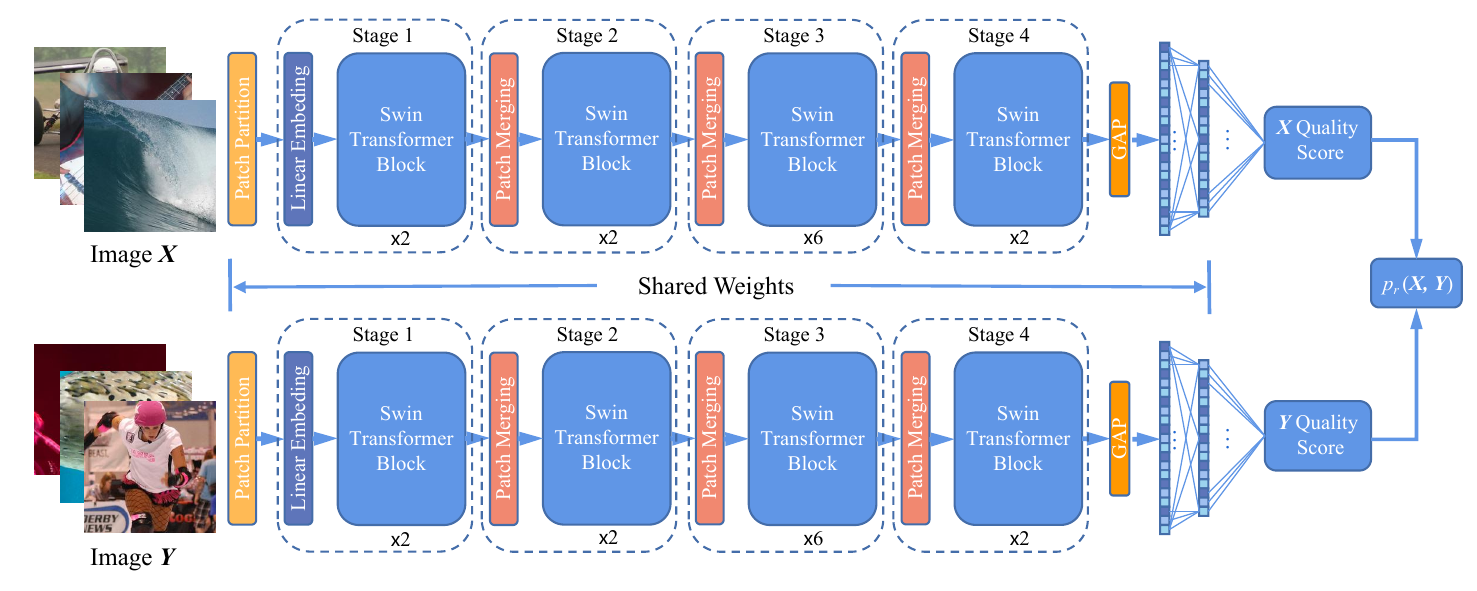}
\caption{Architecture of the CDR-BIQA model. Similar to Siamese Network, the pairwise learning-to-rank framework consists of two streams, each stream is the same as SwinT-IQA.}
\label{fig4}
\end{figure*}

The basic architecture of final CDR-BIQA is the same as SwinT-IQA. But different to SwinT-IQA, the CDR-BIQA is trained with random image pairs $(\boldsymbol{X}, \boldsymbol{Y})$ and corresponding pseudo-label $p_{r}(\boldsymbol{X}, \boldsymbol{Y})$. Therefore, a pairwise learning-to-rank framework consisting of two same streams is adopted, as illustrated in Fig.\ref{fig4}. Although the cross-entropy is widely used for the goal of probability estimation, the cross-entropy loss is unbounded, which may over-penalize some hard training examples, biasing the learned models \cite{Zhang2021UncertaintyAwareBI}. Therefore, the CDR-BIQA is optimized with the fidelity loss $\mathcal{L}_{fidelity }$ \cite{Zhang2021UncertaintyAwareBI},\cite{Tsai2007FRankAR}, which is calculated as follows,

\begin{equation}
\begin{split}
\mathcal{L}_{\textit{fidelity }}=\frac{1}{N_{s}} \sum_{k=1}^{N_{s}}\left(1-\sqrt{\widehat{p_{r k}}(\boldsymbol{X}, \boldsymbol{Y}) p_{r k}(\boldsymbol{X}, \boldsymbol{Y})}\right.\\\left.-\sqrt{\left(1-\widehat{p_{r k}}(\boldsymbol{X}, \boldsymbol{Y})\right)\left(1-p_{r k}(\boldsymbol{X}, \boldsymbol{Y})\right)}\right) ,
\end{split}
\end{equation}
where $p_{r k}(\boldsymbol{X}, \boldsymbol{Y})$ and $\widehat{p_{r k}}(\boldsymbol{X}, \boldsymbol{Y})$ denote the estimated relative probability value and the corresponding pseudo-label of the \textit{k-th} training image pair respectively, and $N_{s}$ is the total number of training pairs. In this paper, we use 500,000 to 1,000,000 random image pairs to train the CDR-BIQA model.

\section{Experiment}

\subsection{Training and Testing Datasets}
In our experiments, five commonly used BIQA datasets are selected for training individual SwinT-IQA models, i.e., LIVEC \cite{Ghadiyaram2016MassiveOC}, BID \cite{Ciancio2011NoReferenceBA}, KonIQ-10k \cite{Hosu2020KonIQ10kAE}, SPAQ \cite{Fang2020PerceptualQA} and FLIVE \cite{Ying2020FromPT}. These datasets are briefly introduced in the following. 

LIVEC \cite{Ghadiyaram2016MassiveOC} consists of 1,162 images with diverse authentic distortions captured by various of mobile devices. BID \cite{Ciancio2011NoReferenceBA} contains 586 images with realistic blur distortion such as out-of-focus, simple motion, complex motion blur, etc. KonIQ-10k \cite{Hosu2020KonIQ10kAE} is composed of 10,073 images which are selected from the large public multimedia database YFCC100m \cite{Thomee2016YFCC100MTN}. The selected images cover a wide and uniform range of distortions in terms of quality indicators such as brightness, colorfulness, contrast, noisiness, sharpness, etc. SPAQ \cite{Fang2020PerceptualQA} consists of 11,125 images taken by 66 kinds of mobile devices. FLIVE \cite{Ying2020FromPT} contains about 40,000 images with authentic distortions and 120,000 randomly cropped patches.

To conduct the large-scale training set for the proposed CDR-BIQA, we used the gMAD in-the-wild dataset \cite{Wang2021LearningFS}, which contains 100,000 real-world images with various scenarios. First, we resize the short side of the different images to 384 while maintaining the pixel aspect-ratio and then crop 224×224 central patches. Afterwards, a total of 500,000 to 1,000,000 pairs of training samples are randomly chosen from this dataset to conduct the large-scale training set. The pseudo-labels are then computed by the proposed strategy. 

\subsection{Implementation Details}
	We use Swin-Transformer-Tiny \cite{Liu2021SwinTH} as the backbone, which has been pre-trained on ImageNet \cite{Deng2009ImageNetAL} to obtain strong feature extraction capability. When training SwinT-IQA on each dataset, we randomly sample and horizontally flipping 10 patches with size 224×224 pixels from each training image for augmentation as in \cite{Kim2017DeepCN}. AdamW \cite{Kingma2015AdamAM} optimizer is adopted with a weight decay of 5×10$^{-4}$. The learning rate is set to 2×10$^{-5}$ and batch size is 32. We use the cosine decay learning rate with the minimum learning rate of 10$^{-8}$, and use linear preheating in first 2 epochs with start learning rate 5×10$^{-7}$. In the test phase, ten patches with 224×224 pixels are randomly cropped from each test image, and the final quality score of an image is the average score of all patches as in \cite{Kim2017DeepCN}. The proposed model is implemented by PyTorch platform with NVIDIA RTX3070 GPU\footnote{The codes and trained model of the CDR-BIQA will be made public.}.

\begin{table*}
\centering  

\renewcommand\arraystretch{1.2}

\fontsize{8}{9}\selectfont 
\caption{Performance evaluation on different BIQA datasets.} 
\begin{tabular}{lcccccccccc}
\toprule[1pt]
\multicolumn{1}{c}{\multirow{2}{*}{}} & \multicolumn{2}{c}{LIVEC}       & \multicolumn{2}{c}{BID}         & \multicolumn{2}{c}{KonIQ-10k}   & \multicolumn{2}{c}{SPAQ}        & \multicolumn{2}{c}{FLIVE}       \\ \cline{2-11} 
\multicolumn{1}{c}{}                  & SRCC           & PLCC           & SRCC           & PLCC           & SRCC           & PLCC           & SRCC           & PLCC           & SRCC           & PLCC           \\ \midrule[1pt]
NIQE                                  & 0.454          & 0.468          & 0.477          & 0.471          & 0.526          & 0.475          & 0.697          & 0.685          & 0.105          & 0.141          \\
ILNIQE                                & 0.453          & 0.511          & 0.495          & 0.454          & 0.503          & 0.496          & 0.719          & 0.654          & 0.219          & 0.255          \\
QAC                                   & 0.069          & 0.014          & 0.326          & 0.323          & 0.343          & 0.296          & 0.047          & 0.107          & 0.104          & 0.066          \\
BMPRI                                 & 0.487          & 0.523          & 0.515          & 0.458          & 0.656          & 0.655          & 0.750          & 0.754          & 0.274          & 0.315          \\
BRISQUE                               & 0.601          & 0.621          & 0.574          & 0.540          & 0.715          & 0.702          & 0.802          & 0.506          & 0.320          & 0.356          \\
CNNIQA                                & 0.627          & 0.601          & 0.616          & 0.614          & 0.685          & 0.684          & 0.796          & 0.799          & 0.306          & 0.285          \\
WaDIQaM-NR                            & 0.692          & 0.730          & 0.653          & 0.636          & 0.729          & 0.754          & 0.840          & 0.845          & 0.435          & 0.430          \\
UNIQUE         & 0.854          & 0.890    & 0.858          & 0.873          & 0.896          & 0.901          & -              & -              & -              & -              \\
DBCNN                                 & 0.844          & 0.862          & 0.845          & 0.859          & 0.878          & 0.887          & 0.910          & 0.913          & {\ul 0.551}    & 0.545          \\
MUSIQ          & 0.702          & 0.746          & -              & -              & 0.916      &0.928    & 0.918          & 0.921          & -              & -  \\
HyperIQA       & 0.855    & 0.871          & \textbf{0.869} & {\ul 0.878}    & 0.908          & 0.921          & 0.916          & 0.919    & 0.535          & 0.623          \\
TReS           & 0.846          & 0.877          & 0.855          & 0.871     & 0.915    & 0.928   & 0.917    & 0.913        & \textbf{0.554} & {\ul0.625} \\
DEIQT          & {\ul 0.875}    & {\ul 0.894}    & -              & -              & {\ul 0.921}    & {\ul 0.934}         & {\ul 0.919}    & {\ul 0.923}    & -              & -  \\
\textbf{SwinT-IQA}                             & \textbf{0.880} & \textbf{0.895} & {\ul 0.864}    & \textbf{0.899} & \textbf{0.932} & \textbf{0.942} & \textbf{0.922} & \textbf{0.925} & 0.547          & \textbf{0.636} \\ \bottomrule[1pt]
\end{tabular}
\label{tab2}
\end{table*}

\subsection{Evaluation Criteria}

We adopted two common performance criterias, i.e., Spearman Rank Correlation Coefficient (SRCC) and Pearson Linear Correlation Coefficient (PLCC). The definitions of SRCC and PLCC are as follows:

\begin{equation}
S R C C=1-\frac{6 \sum_{i=1}^{N} d_{i}^{2}}{N\left(N^{2}-1\right)},
\end{equation}

\begin{equation}
P L C C=\frac{\sum_{i=1}^{N}\left(p_{i}-\bar{p}\right)\left(s_{i}-\bar{s}\right)}{\sqrt{\sum_{i=1}^{N}\left(p_{i}-\bar{p}\right)^{2}\left(s_{i}-\bar{s}\right)^{2}}},
\end{equation}
where $d_{i}$ represents the difference between the ranks of the \textit{i-th} image in subjective and objective assessments, $N$ denotes the number of test images, $s_{i}$ and $p_{i}$ are the MOS and the corresponding objective score of the \textit{i-th} image. $\bar{s}$ and $\bar{p}$ denote the average value of all $s_{i}$ and $p_{i}$. The values of both SRCC and PLCC range from 0 to 1, and higher value represents better performance.

As stated in the report of the video quality experts group (VQEG) \cite{video2003final}, before calculating PLCC, logistic regression is first applied to remove the non-linear ratings caused by human visual observation. The normalized score $\tilde{s}$ is calculated by the following non-linear logic mapping function:

\begin{equation}
\widetilde{s}=\beta_{1}\left(\frac{1}{2}-\frac{1}{\exp \left(\beta_{2}\left(\hat{s}-\beta_{3}\right)\right)}\right)+\beta_{4} \hat{s}+\beta_{5} ,
\end{equation}
where $\beta_{i}(i=1,2,3,4,5)$ denote the regression parameters to be fitted, and $\widehat{s}$ represents the predicted quality score. When training on individual dataset, the dataset is split into a training set of 80\% distorted images and a testing set of 20\% distorted images. In order to reduce the bias caused by the random division between training and testing set, we repeated this process 10 times and reported the median SRCC and PLCC results.

\subsection{Experimental Results}

The proposed model is compared with some typical and traditional BIQA models, such as QAC \cite{Xue2013LearningWH}, NIQE \cite{Mittal2013MakingA}, ILNIQE \cite{Zhang2015AFC}, BRISQUE \cite{Mittal2012NoReferenceIQ} and BMPRI \cite{Min2018BlindIQ}, and several recent SOTA deep BIQA models, i.e., dipIQ \cite{Ma2017dipIQBI}, LFMA \cite{Ma2019BlindIQ}, CNNIQA \cite{Kang2014ConvolutionalNN}, WaDIQaM-NR \cite{Bosse2018DeepNN}, DBCNN \cite{Zhang2020BlindIQ}, HyperIQA \cite{Su2020BlindlyAI}, 
TReS \cite{Golestaneh2022NoReferenceIQ}, PaQ-2-PiQ \cite{Ying2020FromPT}, KonCept512 \cite{Hosu2020KonIQ10kAE}, TRIQ \cite{You2021TransformerFI}, UNIQUE \cite{Zhang2021UncertaintyAwareBI}, MUSIQ\cite{Ke2021MUSIQMI} and DEIQT\cite{Qin2023DataEfficientIQ}. Due to the particularity of the training methods of PaQ-2-PiQ, KonCept512 and TRIQ, we directly use the trained models published by the authors, and only verify their performance by cross-dataset test.

\subsubsection{BIQA Results of SwinT-IQA on Individual Dataset}
To verify the performance and learning ability of the SwinT-IQA model, we have compared it with other BIQA methods on five individual datasets. The BIQA networks are trained on each dataset and then tested on the same dataset. 

\begin{table*}[tp]
\centering  
\renewcommand\arraystretch{1.2}
\fontsize{8}{9}\selectfont 
\caption{SRCC and PLCC results of cross-dataset testing.} 
\begin{tabular}{lccccccccccc}
\toprule[1pt]
\multirow{3}{*}{} & \multirow{3}{*}{Training Set}           & \multicolumn{10}{c}{Testing Set}                                                                                                                                        \\ \cline{3-12} 
                  &                                         & \multicolumn{2}{c}{LIVEC}       & \multicolumn{2}{c}{BID}         & \multicolumn{2}{c}{KonIQ-10k}   & \multicolumn{2}{c}{SPAQ}        & \multicolumn{2}{c}{FLIVE}       \\ \cline{3-12} 
                  &                                         & SRCC           & PLCC           & SRCC           & PLCC           & SRCC           & PLCC           & SRCC           & PLCC           & SRCC           & PLCC           \\ \midrule[1pt]
NIQE              & -                                       & 0.464          & 0.515          & 0.468          & 0.461          & 0.521          & 0.529          & 0.703          & 0.712          & 0.211          & 0.288          \\
ILNIQE            & -                                       & 0.469          & 0.536          & 0.516          & 0.533          & 0.507          & 0.534          & 0.714          & 0.721          & 0.219          & 0.255          \\
dipIQ             & -                                       & 0.187          & 0.290          & 0.009          & 0.346          & 0.228          & 0.437          & 0.385          & 0.497          & 0.088          & 0.053          \\
LFMA              & -                                       & 0.348          & 0.400          & 0.316          & 0.348          & 0.365          & 0.416          & 0.379          & 0.391          & -              & -              \\
DBCNN             & LIVEC                                   & 0.851          & 0.869          & 0.809          & 0.832          & 0.754          & 0.825          & 0.850          & 0.867          & 0.363          & 0.485          \\
DBCNN             & KonIQ-10k                               & 0.755          & 0.777          & 0.815          & 0.818          & 0.875          & 0.884          & 0.850          & 0.862          & 0.399          & 0.509          \\
DBCNN             & BID                                     & 0.751          & 0.814          & 0.845          & 0.859          & 0.734          & 0.795          & 0.820          & 0.848          & 0.319          & 0.417          \\
HyperIQA          & LIVEC                                   & 0.859          & 0.882          & 0.756          & 0.790          & 0.747          & 0.808          & 0.854          & 0.844          & 0.371          & 0.492          \\
HyperIQA          & KonIQ-10k                               & 0.767          & 0.788          & 0.797          & 0.799          & 0.906          & 0.917          & 0.851          & 0.860          & 0.381          & 0.492          \\
HyperIQA          & BID                                     & 0.790          & 0.840          & {\ul 0.869}    & 0.878          & 0.717          & 0.785          & 0.826          & 0.834          & 0.289          & 0.380          \\
TReS              & LIVEC                                   & 0.859          & 0.882          & 0.809          & 0.793          & 0.779          & 0.815          & {\ul 0.864}    & 0.867          & 0.348          & 0.478          \\
TReS              & KonIQ-10k                               & 0.734          & 0.806          & 0.742          & 0.791          & 0.915          & 0.928          & 0.863          & {\ul 0.872}    & 0.346          & 0.474          \\
TReS              & BID                                     & 0.810          & 0.822          & 0.855          & 0.871          & 0.806          & 0.805          & 0.842          & 0.848          & 0.334          & 0.419          \\
PaQ-2-PiQ         & FLIVE                                   & 0.719          & 0.778          & 0.682          & 0.713          & 0.722          & 0.735          & 0.785          & 0.820          & \textbf{0.601} & \textbf{0.685} \\
KonCept512        & KonIQ-10k                               & 0.781          & 0.844          & 0.800          & 0.818          & 0.917    &  0.931   & 0.825          & 0.817          & 0.365          & 0.427          \\
TRIQ              & \multicolumn{1}{l}{LIVEC+KonIQ-10k}     & 0.812          & 0.826          & 0.812          & 0.832          & 0.909          & 0.923          & 0.857          & 0.848          & -              & -              \\
DEIQT             & LIVEC          & 0.875          & 0.894             & -              & -              & 0.744             & -              & -              & -              & -              & -     \\
DEIQT             & KonIQ-10k      & 0.794          & -              & -              & -              & {\ul 0.921}           & {\ul 0.934}              & -              & -              & -              & -     \\
\midrule[1pt]
\textbf{SwinT-IQA}         & LIVEC                                   & {\ul 0.880}    & {\ul 0.895}    & 0.820          & 0.842          & 0.721          & 0.798          & 0.856          & 0.860          & 0.362          & 0.484          \\
\textbf{SwinT-IQA}         & KonIQ-10k                               & 0.816          & 0.832          & 0.843          & 0.835          & \textbf{0.932} & \textbf{0.942} & 0.856          & 0.846          & 0.407          & 0.490          \\
\textbf{SwinT-IQA}         & BID                                     & 0.764          & 0.832          & 0.864          & \textbf{0.899} & 0.689          & 0.732          & 0.817          & 0.839          & 0.292          & 0.406          \\
\textbf{CDR-BIQA}      & \multicolumn{1}{l}{Large-scale Dataset} & \textbf{0.888} & \textbf{0.906} & \textbf{0.881} & {\ul 0.891}    & 0.909          & 0.922          & \textbf{0.904} & \textbf{0.901} & {\ul 0.474}    & {\ul 0.577}    \\ \bottomrule[1pt]
\end{tabular}
\label{tab3}
\end{table*}

Table \ref{tab2} lists the BIQA results on each dataset. From the table, we can observe that deep BIQA models can outperform traditional methods, because these five datasets are all composed of in-the-wild images, and thus are difficult for traditional BIQA methods without specific training. By comparing deep BIQA networks, the SwinT-IQA can achieve the best performance. It proves that the SwinT backbone can have better learning ability than other SOTA models for BIQA task. 

In addition, we can find that many methods do not perform as well as other datasets on the FLIVE dataset, because the MOS distribution of the FLIVE dataset is mainly concentrated on high-quality scores (about 75), while the MOS distribution of other dataset covers more evenly from low-quality scores to high-quality scores. The difference of score distribution of datasets leads to this problem.

\subsubsection{Results of Cross-Dataset Testing}
Results on individual dataset can verify the learning ability of different models, but satisfactory performance on specific dataset does not mean high generalization capability for real-world images outside of the dataset. Due to the influence of different cameras, shooting environment, photography skills and other factors, in practical applications, the content and distortion of an image may be different from that of the training datasets. Therefore, cross-dataset evaluation is very important for the BIQA task, which can reflect the generalization ability of an algorithm for images obtained in completely different ways. Table \ref{tab3} shows the results of cross-dataset testing, from which we can get the following findings. First, the proposed CDR-BIQA can achieve the best performance on most of the testing datasets. Second, on KonIQ-10k dataset, the CDR-BIQA is merely worse than TReS, SwinT-IQA and DEIQT trained directly on this dataset. Similar findings can be found in FLIVE dataset. CDR-BIQA outperforms other methods except the PaQ-2-PiQ trained on FLIVE. Third, interesting observations can be obtained from LIVEC and BID datasets. CDR-BIQA can achieve the best performance, even better than original SwinT-IQA models learned directly on these datasets. In addition, CDR-BIQA can also perform the best on SPAQ dataset. Note that the CDR-BIQA is trained on the proposed large-scale dataset instead of these datasets used for testing, which means images in the training and testing dataset are from different scenarios. Therefore, the cross-dataset testing results demonstrate the generalization ability of the proposed method.

\begin{figure*}
\centering
\includegraphics[height=9.5cm]{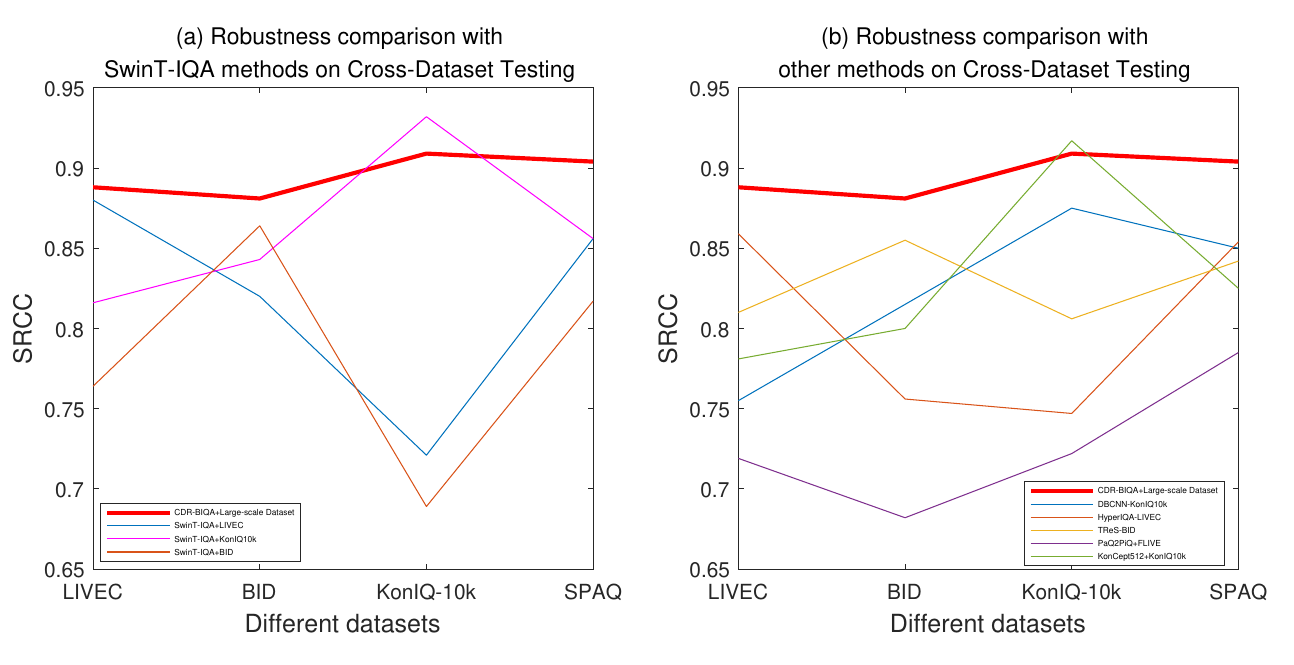}
\caption{Robustness comparison on Cross-Dataset Testing. The other IQA models only achieve peak performance when they are evaluated on datasets matching their training sets, with performance declining on other testing sets. In contrast, the proposed method can achieve stronger cross-dataset robustness.} 
\label{fig6}
\end{figure*}

Furthermore, to more intuitively compare the robustness of each method on cross-dataset testing, we depict the results using line graphs in Fig. \ref{fig6}. The legend indicates IQA model and corresponding training set, while the horizontal axis represents the different testing sets used for evaluation. It is obvious that these comparative methods exhibit substantial performance variation across different datasets, with models trained on the matching training set achieving the highest SRCC values. Only our proposed method 
achieves consistently good performance across all testing sets, demonstrating the cross-dataset robustness of the proposed approach.

\subsection{Ablation Study}
\subsubsection{Effects of Different Pseudo-Labels}
We have studied the impact of pseudo-labels generated by different combination of SwinT-IQA models. From Table \ref{tab4}, it can be observed that, in most cases, as the number of datasets increases, the robustness of CDR-BIQA has been improved. However, for the BID dataset, because of its small scale, it is not helpful for training a relatively robust BIQA. Furthermore, even if we did not use the SwinT-IQA trained on the BID dataset to calculate pseudo-labels, the SRCC and PLCC results of final CDR-BIQA are higher than the SwinT-IQA trained on the BID dataset. Hence, in this paper, the default pseudo-labels are computed via SwinT-IQA models trained on four datasets, i.e., LIVEC, KonIQ-10k, SPAQ and FLIVE.

\begin{table*}
\centering  
\renewcommand\arraystretch{1.2}
\fontsize{8}{9}\selectfont 
\caption{SRCC and PLCC results of CDR-BIQA trained with different pseudo-label selection.} 
\begin{tabular}{lcccccccccc}
\toprule[1pt]
\multirow{3}{*}{SwinT-IQA Trained on} & \multicolumn{10}{c}{Testing Set}                                                                                                                                        \\ \cline{2-11} 
                                             & \multicolumn{2}{c}{LIVEC}       & \multicolumn{2}{c}{BID}         & \multicolumn{2}{c}{KonIQ-10k}   & \multicolumn{2}{c}{SPAQ}        & \multicolumn{2}{c}{FLIVE}       \\ \cline{2-11} 
                                             & SRCC           & PLCC           & SRCC           & PLCC           & SRCC           & PLCC           & SRCC           & PLCC           & SRCC           & PLCC           \\ \midrule[1pt]
LIVEC+KonIQ-10k                              & \textbf{0.892} & \textbf{0.908} & 0.876          & 0.888          & \textbf{0.910} & \textbf{0.924} & 0.874          & 0.872          & 0.424          & 0.525          \\ 
LIVEC+KonIQ-10k+SPAQ                         & 0.885          & {\ul 0.906}    & {\ul 0.884}    & {\ul 0.896}    & 0.906          & {\ul 0.923}    & \textbf{0.907} & \textbf{0.903} & {\ul 0.439}    & {\ul 0.541}    \\ 
LIVEC+KonIQ-10k+SPAQ+BID                     & 0.886          & {\ul 0.906}    & \textbf{0.907} & \textbf{0.917} & 0.888          & 0.907          & 0.898          & 0.892          & 0.429          & 0.527          \\ 
LIVEC+KonIQ-10k+SPAQ+FLIVE                   & {\ul 0.888}    & {\ul 0.906}    & 0.881          & 0.891          & {\ul 0.909}    & 0.922          & {\ul 0.904}    & {\ul 0.901}    & \textbf{0.474} & \textbf{0.577} \\ \bottomrule[1pt]
\end{tabular}
\label{tab4}
\end{table*}

\subsubsection{Extended Cross-Dataset Testing on synthetically distorted datasets.}

Limited by high cost, there is few BIQA datasets with MOS labels at present. Hence, we also implement extended cross-dataset test on synthetically distorted datasets, as listed in Table \ref{tab5}. Although distortions in synthetic datasets are quite different to real-world distortions in BIQA datasets, the proposed CDR-BIQA still achieves better testing results than other BIQA networks. This also partially demonstrates the robustness of the proposed method.

\begin{table*}[]
\centering  
\renewcommand\arraystretch{1.2}
\fontsize{7.5}{9}\selectfont 
\caption{Extended cross-dataset testing results of different BIQA methods on synthetically distorted datasets.} 
\begin{tabular}{lccrcr}
\toprule[1pt]
\multicolumn{1}{c}{\multirow{3}{*}{}} & \multirow{3}{*}{Training Set} & \multicolumn{4}{c}{Testing Set}                                                       \\ \cline{3-6} 
\multicolumn{1}{c}{}                  &                               & \multicolumn{2}{c}{LIVE}                  & \multicolumn{2}{c}{CSIQ}                  \\ \cline{3-6} 
\multicolumn{1}{c}{}                  &                               & SRCC           & \multicolumn{1}{c}{PLCC} & SRCC           & \multicolumn{1}{c}{PLCC} \\ \midrule[1pt]
DBCNN                                 & KonIQ-10k                     & 0.787          & 0.767                    & 0.509          & 0.500                    \\
HyperIQA                              & KonIQ-10k                     & 0.758          & 0.748                    & 0.529          & 0.549                    \\
TReS                                  & KonIQ-10k                     & 0.761          & 0.744                    & 0.570          & 0.580                    \\
KonCept512                            & KonIQ-10k                     & 0.777          & 0.745                    & {\ul 0.676}    & 0.672                    \\
PiQ-2-PaQ                             & FLIVE                         & 0.563          & 0.542                    & 0.576          & {\ul 0.710}              \\
TRIQ                                  & LIVEC+KonIQ-10k               & 0.658          & 0.704                   & 0.512          & 0.599                    \\ \midrule[1pt]
\textbf{SwinT-IQA}                    & KonIQ-10k                     & {\ul 0.851}    & {\ul 0.852}              & 0.674          & 0.708                    \\
\textbf{CDR-BIQA}                     & Large-scale Dataset           & \textbf{0.894} & \textbf{0.889}           & \textbf{0.701} & \textbf{0.741}           \\  \bottomrule[1pt]
\end{tabular}
\label{tab5}
\end{table*}

\subsubsection{Effects of Different Numbers of Training Pairs}
Table \ref{tab6} shows the results of different numbers of training pairs. With the increase of training image pairs, the performance of CDR-BIQA is gradually improved. This indicates that more training samples can refine the BIQA performance. However, when the image pairs exceed 500,000, the performance is not significantly improved. Moreover, too many training samples naturally take more training time. By balancing performance and training time, it is a relatively better choice to use 500,000 training pairs.
\begin{table*}[]
\centering  
\renewcommand\arraystretch{1.2}
\fontsize{8}{9}\selectfont 
\caption{SRCC and PLCC results of CDR-BIQA trained with different numbers of image pairs.} 
\begin{tabular}{ccccccccccc}
\toprule[1pt]
\multirow{3}{*}{Number of Pairs} & \multicolumn{10}{c}{Testint Set}                                                                                                                                               \\ \cline{2-11} 
                                 & \multicolumn{2}{c}{LIVEC}       & \multicolumn{2}{c}{BID}         & \multicolumn{2}{c}{KonIQ-10k}   & \multicolumn{2}{c}{SPAQ}        & \multicolumn{2}{c}{FLIVE}       \\ \cline{2-11} 
                                 & SRCC           & PLCC           & SRCC           & PLCC           & SRCC           & PLCC           & SRCC           & PLCC           & SRCC           & PLCC           \\ \midrule[1pt]
50,000                           & 0.874          & 0.886          & 0.875          & 0.881          & 0.897          & 0.912          & {\ul 0.901}    & 0.898          & 0.461          & 0.564          \\ 
100,000                          & {\ul 0.878}    & 0.897          & 0.876          & {\ul 0.888}    & 0.902          & 0.919          & {\ul 0.901}    & {\ul 0.901}    & 0.471          & {\ul 0.574}    \\ 
200,000                          & \textbf{0.888} & 0.905          & {\ul 0.877}    & \textbf{0.891} & {\ul 0.907}    & {\ul 0.922}    & \textbf{0.904} & \textbf{0.903} & {\ul 0.472}    & 0.573          \\ 
500,000                          & \textbf{0.888} & {\ul 0.906}    & \textbf{0.881} & \textbf{0.891} & \textbf{0.909} & {\ul 0.922}    & \textbf{0.904} & {\ul 0.901}    & \textbf{0.474} & \textbf{0.577} \\ 
1,000,000                        & \textbf{0.888} & \textbf{0.908} & 0.875          & {\ul 0.888}    & \textbf{0.909} & \textbf{0.923} & \textbf{0.904} & 0.898          & \textbf{0.474} & \textbf{0.577} \\ \bottomrule[1pt]
\end{tabular}
\label{tab6}
\end{table*}

\begin{table*}[]
\centering  
\renewcommand\arraystretch{1.2}
\fontsize{8}{9}\selectfont 
\caption{SRCC and PLCC results of training and testing on a single dataset with different backbones.}
\begin{tabular}{ccccccccccc}
\toprule[1pt]
\multirow{2}{*}{Backbone} & \multicolumn{2}{c}{LIVEC}       & \multicolumn{2}{c}{BID}         & \multicolumn{2}{c}{KonIQ-10k}   & \multicolumn{2}{c}{SPAQ}        & \multicolumn{2}{c}{FLIVE}       \\ \cline{2-11} 
                          & SRCC           & PLCC           & SRCC           & PLCC           & SRCC           & PLCC           & SRCC           & PLCC           & SRCC           & PLCC           \\ \midrule[1pt]
ResNet-50                 & 0.857          & 0.876          & 0.842          & 0.854          & 0.907          & 0.912          & 0.912          & 0.916          & 0.511          & 0.490          \\
ConvNeXt-Tiny             & 0.852          & 0.884          & 0.841          & 0.874          & 0.926          & 0.939          & 0.920          & 0.924          & 0.538          & 0.629          \\
Swin-Transformer-Tiny     & \textbf{0.880} & \textbf{0.895} & \textbf{0.864} & \textbf{0.899} & \textbf{0.932} & \textbf{0.942} & \textbf{0.922} & \textbf{0.925} & \textbf{0.547} & \textbf{0.636} \\ \bottomrule[1pt]
\end{tabular}
\label{tab7}
\end{table*}

\begin{table*}[]
\centering  
\renewcommand\arraystretch{1.2}
\fontsize{8}{9}\selectfont 
\caption{SRCC and PLCC results of cross-dataset testing with different backbones.} 
\begin{tabular}{cccccccccccc}
\toprule[1pt]
\multirow{3}{*}{Backbone} & \multirow{3}{*}{Training Set}        & \multicolumn{10}{c}{Testing Set}                                                                                                                                        \\ \cline{3-12} 
                          &                                      & \multicolumn{2}{c}{LIVEC}       & \multicolumn{2}{c}{BID}         & \multicolumn{2}{c}{KonIQ-10k}   & \multicolumn{2}{c}{SPAQ}        & \multicolumn{2}{c}{FLIVE}       \\ \cline{3-12} 
                          &                                      & SRCC           & PLCC           & SRCC           & PLCC           & SRCC           & PLCC           & SRCC           & PLCC           & SRCC           & PLCC           \\ \midrule[1pt]
ResNet-50                 & \multirow{3}{*}{\begin{tabular}[c]{@{}c@{}}Large-scale\\  Dataset\end{tabular}} & 0.862          & 0.865          & 0.839          & 0.834          & 0.872          & 0.884          & 0.898          & 0.876          & 0.444          & 0.532          \\
ConvNeXt-Tiny             &                                      & \textbf{0.890} & 0.903          & 0.857          & 0.871          & 0.887          & 0.908          & 0.892          & 0.882          & \textbf{0.493} & \textbf{0.591} \\
Swin-Transformer-Tiny     &                                      & 0.888          & \textbf{0.906} & \textbf{0.881} & \textbf{0.891} & \textbf{0.909} & \textbf{0.922} & \textbf{0.904} & \textbf{0.901} & 0.474          & 0.577          \\ \bottomrule[1pt]

\end{tabular}
\label{tab8}
\end{table*}

\subsubsection{Effects of Different Backbones}
We conduct experiments by changing the backbone of the model to verify the effectiveness of the proposed learning strategy. For a fair comparison, we choose pretrained ResNet-50 \cite{He2016DeepRL}, ConvNeXt-Tiny \cite{liu2022convnet} and Swin-Transformer-Tiny \cite{Liu2021SwinTH} in our experiments. Through the results of training and testing on single dataset in Table \ref{tab7}, we can find that Swin-Transformer-Tiny can achieve the best results on single dataset compared to the other two backbones. 

\begin{figure}
\centering
\includegraphics[width=1.0\columnwidth]{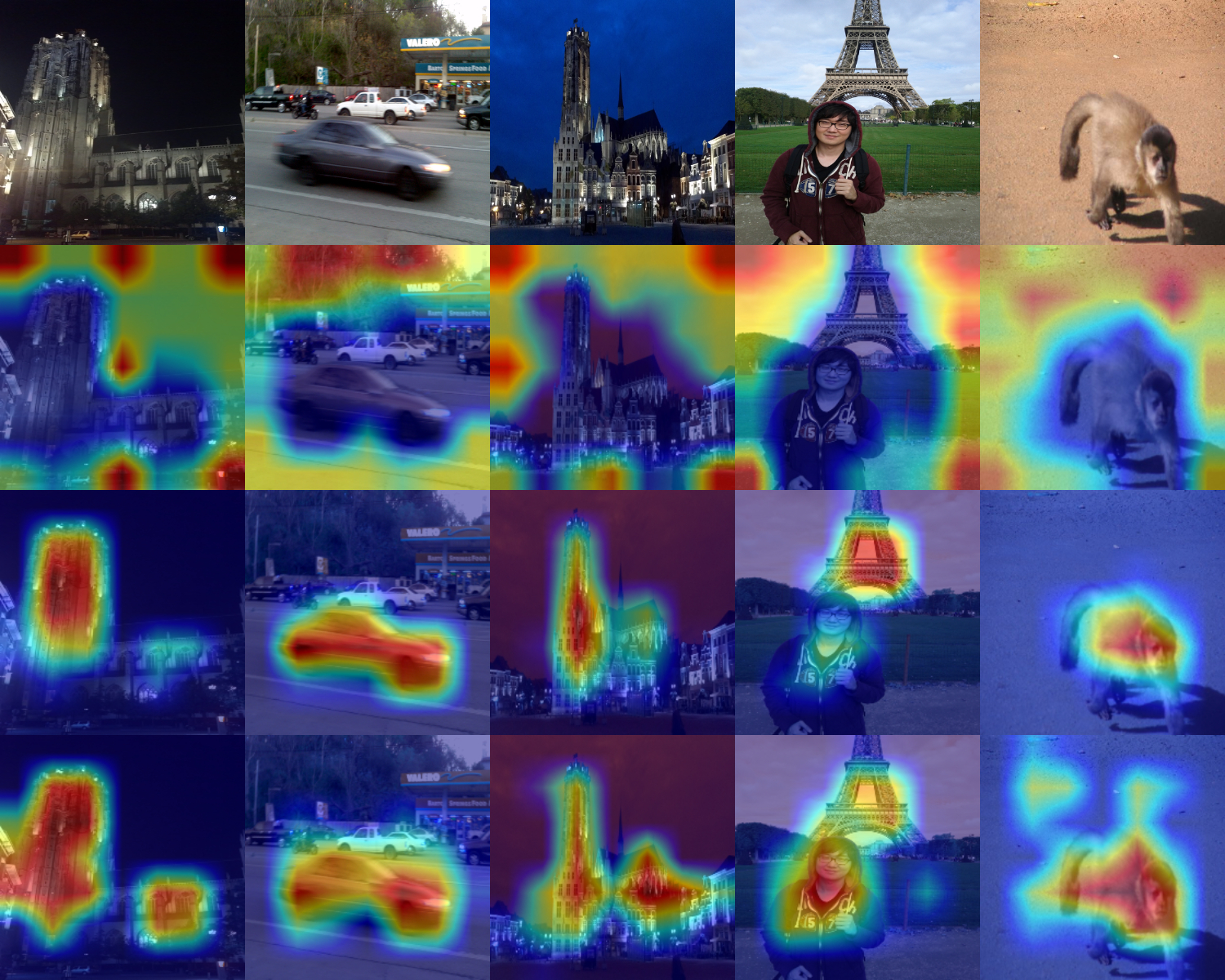}
\caption{Visualization of the last norm layer of network trained on different datasets via GradCAM tools \cite{Selvaraju2017GradCAMVE}. (First row: images in the LIVEC dataset, second row: SwinT-IQA trained on KonIQ-10k dataset, third row: SwinT-IQA trained on LIVEC dataset, final row: CDR-BIQA)}\label{fig5}
\end{figure}

Cross-dataset test results of these backbones are listed in Table \ref{tab8}. We can find that the proposed method with different backbones all perform well in cross-dataset testing, which shows the effectiveness of the proposed training strategy. Meanwhile, the SwinT-based model performs slightly better than other backbones.

\subsubsection{Visualization of Result}
In order to visually demonstrate the cross-dataset robustness of the proposed method, we have used the last normalization layer activations maps by means of Grad-CAM tools \cite{Selvaraju2017GradCAMVE}, which show essential features extracted for IQA by different training settings. The corresponding heat maps and inputs are demonstrated in Fig. \ref{fig5}. As illustrated in Fig. \ref{fig5}, SwinT-IQA trained on LIVEC dataset can perceive important regions in LIVEC images. However, SwinT-IQA trained on KonIQ-10k dataset cannot focus on correct regions in LIVEC images. At last, the CDR-BIQA can perceive reasonable salient regions, which also demonstrate its generalization ability.

\section{Conclusion}
In order to improve the generalization ability of blind image quality assessment (BIQA) methods in practical scenarios, this paper proposed a cross-dataset-robust BIQA network and training strategy. First, biased Swin-Transformer-based image quality assessment (SwinT-IQA) models are separately trained on current BIQA datasets with subjective scores. Second, these biased SwinT-IQA models are jointly used to estimate pseudo-labels for a large-scale real-world image dataset. Note that the pseudo-labels use the relative probability of two random images instead of fixed quality scores. At last, the final cross-dataset-robust BIQA model is optimized on the proposed large-scale image dataset with learning-to-rank framework. Experimental results on cross-dataset tests show that the proposed method can achieve better robustness and generalization ability for real-world images than many state-of-the-art methods.

\bibliographystyle{IEEEtran}
\bibliography{Ref}

\end{document}